\documentclass[journal, twoside]{IEEEtran}
\let\labelindent\relax
\usepackage{amsmath}
\usepackage{amssymb}
\usepackage{amsfonts}
\usepackage{mathtools}
\usepackage{latexsym}
\usepackage{tabularx}
\usepackage{colortbl}
\usepackage{ulem}
\usepackage{threeparttable}
\usepackage{booktabs}
\usepackage{graphicx}
\usepackage{wrapfig}
\usepackage[lined, ruled, linesnumbered, commentsnumbered, noend]{algorithm2e}
\usepackage[colorlinks=true, allcolors=blue]{hyperref}

\SetCommentSty{mycommfont}
\usepackage{subcaption}
\usepackage{amsfonts}
\usepackage{enumitem}
\usepackage{lipsum}
\usepackage{stfloats}
\usepackage[numbers]{natbib}
\usepackage[usenames,dvipsnames]{xcolor}
\usepackage{comment}

\usepackage{multirow}

\newcommand{\rev}[1]{\textcolor{black}{#1}}
\usepackage{tikz}
\usetikzlibrary{fit,shapes.misc}

\newcommand{\circnum}[1]{\textcircled{\raisebox{-0.25ex}{#1}}}
\usepackage{CJKutf8}
\definecolor{bleudefrance}{rgb}{0.19, 0.55, 0.91}
\definecolor{awesome}{rgb}{1.0, 0.13, 0.32}

\definecolor{darkgreen}{rgb}{0.0, 0.65, 0.0}
\definecolor{babyblue}{rgb}{0.29, 0.75, 0.93}

\begin{document}
\title{Bimanual Regrasp Planning and Control for Active Reduction of Object Pose Uncertainty}
\author{Ryuta Nagahama$^{1}$, Weiwei Wan$^{*1}$, Zhengtao Hu$^{2}$, Kensuke Harada$^{1}$
\thanks{Manuscript received: Feb 28th, 2025; Revised: May 6th, 2025; Accepted: Jun 2nd, 2025. This paper was recommended for publication by Editor Olivier Stasse upon evaluation of the Associate Editor and Reviewers’ comments. The work was supported by JSPS KAKENHI Grant Number JP22K12205 and partially a joint project with Panasonic Holdings Corporation P0001266. ${}^1$R. Nagahama, W. Wan, and K. Harada are with Graduate School of Engineering Science, Osaka University, Japan. ${}^1$Z. Hu is with School of Mechatronic Engineering and Automation, Shanghai University, China. His contribution was supported by National Natural Science Foundation of China (62403297). ${}^{*}$Contact: Weiwei Wan, {wan.weiwei.es@osaka-u.ac.jp}. Digital Object Identifier (DOI):}}
\markboth{IEEE ROBOTICS AND AUTOMATION LETTERS. PREPRINT VERSION. ACCEPTED JUNE, 2025.}
{Nagahama \MakeLowercase{\textit{et al.}}: Bimanual Regrasp Planning for Reducing Object Pose and Grasping Uncertainty}
\maketitle


\bstctlcite{IEEEexample:BSTcontrol}
\begin{abstract}
Precisely grasping an object is a challenging task due to pose uncertainties. Conventional methods have used cameras and fixtures to reduce object uncertainty. They are effective but require intensive preparation, such as designing jigs based on the object geometry and calibrating cameras with high-precision tools fabricated using lasers. In this study, we propose a method to reduce the uncertainty of the position and orientation of a grasped object without using a fixture or a camera. Our method is based on the concept that the flat finger pads of a parallel gripper can reduce uncertainty along its opening/closing direction through flat surface contact. Three \rev{approximately} orthogonal grasps by parallel grippers with flat finger pads collectively constrain an object's position and orientation to a unique state. Guided by the concepts, we develop a regrasp planning and admittance control approach that sequentially finds and leverages three \rev{approximately} orthogonal grasps of two robotic arms to actively reduce uncertainties in the object pose. We evaluated the proposed method on different initial object uncertainties and verified that it had good repeatability. The deviation levels of the experimental trials were on the same order of magnitude as those of an optical tracking system, demonstrating strong relative inference performance.
\end{abstract}


\section{Introduction} 

\IEEEPARstart{A} significant challenge in robotic manipulation lies in addressing the uncertainties associated with object grasping. The uncertainties often arise from errors in environmental registration, inaccuracies in object pose recognition, and unbalanced contact during grasping that leads to pose deviations. The uncertainties can result in discrepancies between the actual and expected poses of objects or tools, potentially causing task failures. Existing methods used to address uncertainty problems typically fall into two categories: Sensor-based correction methods \cite{du2021vision}\cite{yan2021soft} and fixture-based constraint methods \cite{chavan2015prehensile}\cite{fiedler2024jigs}. Sensor-based methods leverage additional hardware, such as force sensors or tactile sensors, to detect and mitigate uncertainties. Fixture-based methods, on the other hand, rely on structuring the environment or using special-purpose structures to constrain object pose. However, both approaches have notable limitations, as they often require additional hardware adjustments or environment modifications.

\begin{figure}[tbp]
    \centering
    \includegraphics[width=\linewidth]{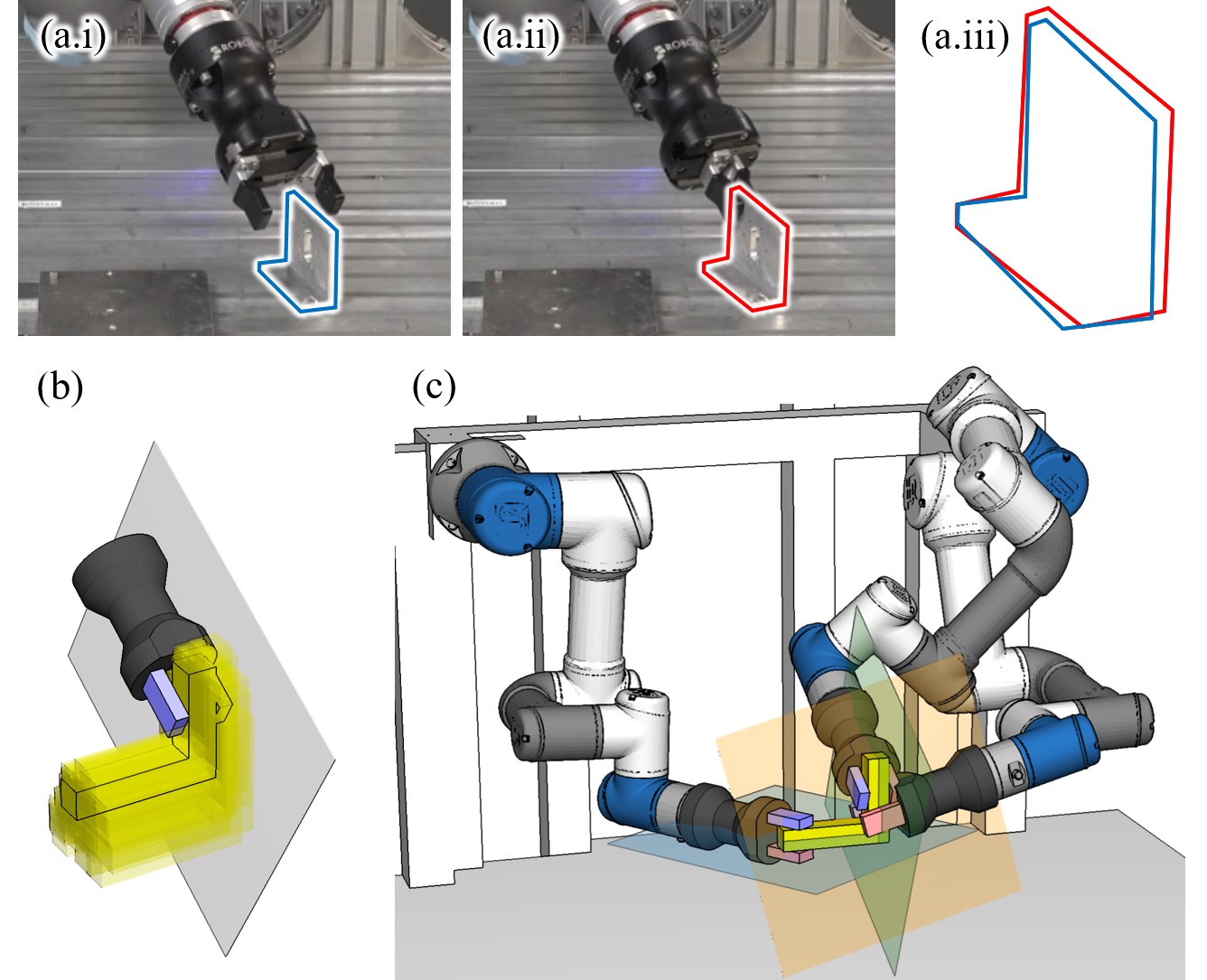}
    \caption{(a) Precisely grasping an object is difficult. (a.i) Before grasping. (a.ii) After grasping. (a.iii) Comparison. (b) Flat finger pads of a parallel gripper help constrain uncertainty to a plane. (c) Three aprpoximately orthogonal parallel grasps constrain three planes, leading to an exact pose.}
    \label{features}
\end{figure}  

This paper proposes a bimanual regrasp planning method that leverages flat finger pad constraints and admittance control to reduce uncertainties. It does not require external fixtures or additional sensors and is more flexible for diverse robotic manipulation scenarios. The fundamental concept of the proposed method is illustrated in Fig. \ref{features}(b) and (c). It is based on the observation that the flat finger pads of parallel grippers can reduce uncertainty along the hand’s opening/closing direction through flat surface contact. Three such grasps collectively constrain the object's position and orientation to a unique state\footnote{This unique pose constraint holds under the following assumptions: i) The object is sufficiently rigid such that the grasping force is transmitted and sensed through the force sensors, rather than causing deformation. ii) The contact between the gripper fingers and the object is assumed to be flat surface contact with sufficient friction to prevent slippage. iii) The maximum uncertainty in the object pose is within the gripper’s opening range, ensuring valid contact during grasping.}. Guided by the concepts, we develop a regrasp planning approach \cite{wan2015improving} that sequentially executes a series of regrasping motions using two robotic arms while selecting \rev{approximately} orthogonal grasp triplets to simultaneously reduce uncertainties in the object pose. However, grasp planning alone does not fully achieve our goal, as simply executing the planned motion sequence using position control may exert excessive grasping forces on the uncertain object and lead to deformation or even damage. To address this issue, we develop an admittance control \cite{mae2024admittance} method that utilizes force/torque sensors at the robots' wrists to adjust the grasping pose and conform to the uncertain object pose. By integrating regrasp planning with admittance control, our method iteratively reduces uncertainties along three \rev{approximately} orthogonal axes, thus achieving an accurate object pose estimation. 

In the experiments, we implemented the proposed method and evaluated its performance on different initial object uncertainties. The results demonstrate that the method achieves satisfactory repeatability. The deviation levels of remaining uncertainties were the same order of magnitude as those of an optical tracking system, demonstrating that the method provides strong relative inference performance. By leveraging the robot’s inherent geometric constraints and force sensors, the proposed approach can help actively reduce uncertainties in the grasped object.

\section{Related Work}

\subsection{Categories of Regrasp Planning Methods}
Regrasp planning aims to automatically generate grasp sequences to accomplish object manipulation. It involves two main categories: discrete and continuous.

Discrete regrasp planning focuses on identifying discrete pick-and-place sequences. For example, Liu et al. \cite{liu2024novelplanningframeworkcomplex} developed regrasp planning using multiple mobile robots. The method achieves flipping manipulation with the help of regrasping. Kim et al. \cite{kim2016randomized} developed a randomized planning method for regrasp planning. Baek et al. \cite{baek2021pre} considered non-prehensile grasping in regrasp planning and secured power grasps for stable holding. Qin et al. \cite{qin2023dual} presented a regrasp planning method for installing a deformable linear object. Rigidity constraints were considered when planning regrasp motion. Mitrano et al. \cite{mitrano2024grasploop} studied a similar problem and proposed considering avoiding environment-grasp-body loops to accelerate grasp selection and planning. Sundaralingam et al. \cite{sundaralingam2018geometric} extended regrasp planning to a multi-fingered hand. Murooka et al. \cite{murooka2021humanoid} proposed the foot-step and regrasping planning method for a humanoid robot to simultaneously determine foot steps and hand switches. 

Continuous regrasp methods, on the other hand, aim to adjust object poses through controlled rolling or sliding motions while maintaining contact. For instance, Maximo et al. \cite{roa2009regrasp} presented the grasp space method for continuous regrasp (rolling or sliding) that maintains force closures. Hu et al. \cite{hu2023multi} formulated regrasp planning as a multi-modal problem, where three manifolds, including the regrasp manifold, the transfer manifold, and the slide manifold, were planned across to determine action sequences and thus plan motion. Cheng et al. \cite{cheng2022contact} further proposed the contact mode method to guide manipulation planning across multiple manifolds. 

The work presented in this paper is based on iterative pick-and-place actions using two robot hands, and thus it falls under the first type of regrasp planning. \rev{It can be seen as a concrete realization of a flipped funnel \cite{matt85}, where the object, through interaction, progressively guides the three structured grasping poses, allowing its uncertain state to be estimated toward a determined pose.}

\subsection{Key Challenges and the Focus of this Work}
One of the most critical challenges in regrasp planning is the planning speed. Researchers have explored various methods to enhance efficiency. For example, Cho et al. \cite{cho2003complete} developed a fast regrasp planning algorithm using look-up tables and two-finger parallel grippers. Chiu et al. \cite{chiu2021bimanual} proposed a reinforcement learning method to accelerate the planning of regrasp poses. Their actions were defined in the local frame of a robot hand, thereby resolving the dependencies on robotic kinematics. Wada et al. \cite{wada2022reorientbot} significantly improved planning speed by using a neural network to filter waypoint poses and quickly select grasp poses. Xu et al. \cite{xu2022efficient} proposed learning reorientation object poses and then planning regrasp motion while considering the learned middle reorientation poses. Learning-based methods have demonstrated effectiveness in improving regrasp planning efficiency. While there are numerous other publications, this paper does not focus heavily on this aspect and thus will not delve into further details.

Another critical challenge in regrasp planning is uncertainty. Repeated regrasping can result in error accumulation, causing the grasp pose to deviate significantly from the planned outcome. The deviation can prevent the robot from successfully executing subsequent tasks. To address the uncertainty problems, Bauza et al. \cite{bauza2024simple} integrated tactile information into regrasp planning and achieved precise and robust pick and placement for regrasping. Bronars et al. \cite{bronars2024texterity} proposed a simultaneous tactile estimation and control method to plan sliding regrasp motion while maintaining grasping precision continuously. Gualtieri et al. \cite{gualtieri2021robotic} estimated the quality of grasp candidates against uncertainty based on point cloud completion and optimized regrasp sequences to be robust. Zhang et al. \cite{zhang2024learning} used reinforcement learning to learn a regrasp policy that optimizes grasp stability. Both vision and visual-tactile information were used for learning. This study can be partially classified as research to reduce uncertainties in grasping poses during the regrasp planning process. However, our primary goal is not to achieve precise pose estimation for regrasping. Instead, we focus on actively leveraging the consecutive geometric constraints inherent in the regrasping process \cite{nikhil14extrinsic}\cite{zhou2018convex}\cite{mannam2019sensorless}\cite{von2022uncertainty} to reduce uncertainty.

From a broader perspective, our research also falls within the category of methods that utilize interactive perception \cite{jeannette17} to estimate the pose of grasped objects. In addition to the grasping-based interaction approaches, recent literature on non-prehensile interaction has been receiving growing attention. For example, Jankowski et al. \cite{jankowski24} discussed the stochastic contact dynamics of pushing and proposed a pushing planning method that minimizes uncertainty considering informed priors. Dong et al. \cite{dong24}, on the other hand, modeled the contact dynamics of pushing as an energy model and used constraints on the energy margin to plan robust pushing.

\section{Regrasp Planning Using Orthogonal Triplets}

\begin{figure}[!htbp]
    \centering
    \includegraphics[width=\linewidth]{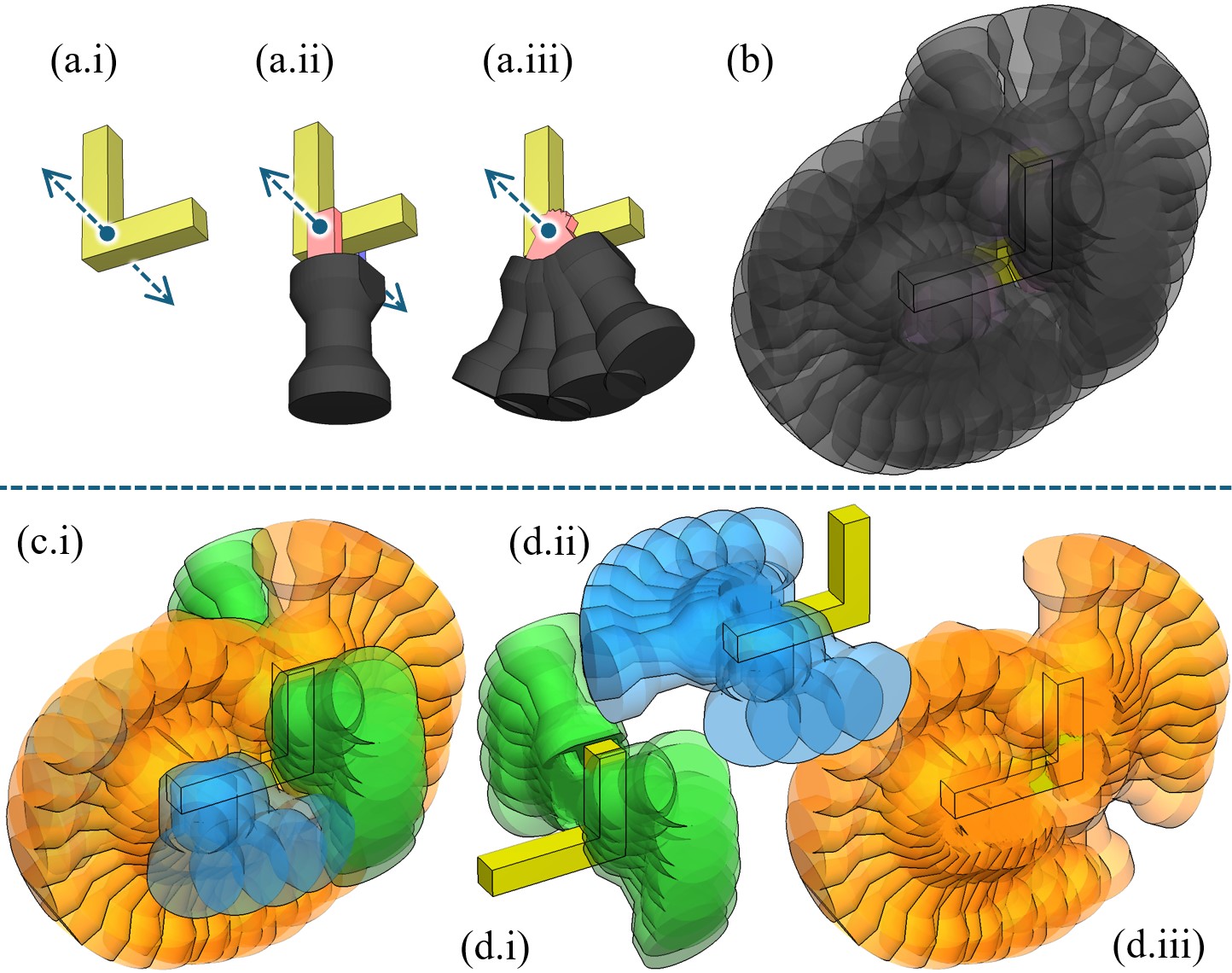}
    \caption{(a) Antipodal grasp planning. (b) Candidate grasp poses planned using the antipodal method. (c) Separating the candidate grasp poses into three orthogonal groups considering their opening and closing directions. Each color in the subfigure represents one group.}
    \label{grasp_planning}
\end{figure}

To achieve the goal of this study, we need to identify grasp poses along three \rev{approximately} orthogonal directions. Given a 3D model of an object, our method begins with antipodal grasp planning \cite{chen1993finding}. During the grasp planning, we sample contact points on the object's surface. For each sampled contact point, a ray is cast in the direction opposite to the surface normal of the point. The intersection between this ray and the object's surface is determined using ray tracing. If the surface normal at the intersection point is opposite in direction to the normal at the ray’s starting point (satisfying the parallel grasp constraint), this intersection point and the starting point are stored as a candidate pair. For all candidate pairs, we align the gripper fingers' contact centers with these points and rotate the gripper about the axis passing through the pair. Gripper poses that do not result in collisions are extracted as candidate grasps. Fig. \ref{grasp_planning}(a.i$\sim$a.iii) illustrates the grasp planning process on a single contact point. The greater the number of sampled contact points, the larger the number of generated candidate grasps. Fig. \ref{grasp_planning}(b) demonstrates the generated candidate grasps when the number of contact points is set to 5.

Subsequently, based on the results of the grasp planning, we separate the grasps into groups. Since our goal is to identify three \rev{approximately} orthogonal grasp poses, the grouping criterion is defined by the opening and closing directions of the gripper fingers. Grasp poses with the same opening and closing direction are grouped together, and each group can be represented by a unique opening and closing vector, denoted as $\boldsymbol{v}_i$\footnote{In this context, the direction of $\boldsymbol{v}_i$ is not considered.}. Through the grouping process, we can obtain several groups represented by $\boldsymbol{v}_1$, $\boldsymbol{v}_2$, \ldots. The candidate grasp poses in Fig. \ref{grasp_planning}(b) are classified into three strictly orthogonal groups. They are visualized in Fig. \ref{grasp_planning}(c) using red, green, and blue.

After obtaining the grasp groups, we perform triplet combinations of the groups to identify three groups that satisfy the \rev{approximately} orthogonal constraint. To evaluate whether the constraint is satisfied, we use the following formula to compute a score for each triplet combination.
\begin{align}
    \mathtt{score} = |\boldsymbol{v}_i \cdot \boldsymbol{v}_j| + |\boldsymbol{v}_i \cdot \boldsymbol{v}_k| + |\boldsymbol{v}_j \cdot \boldsymbol{v}_k|
    \label{eq_score}
\end{align}

The $\boldsymbol{v}_i$, $\boldsymbol{v}_j$, and $\boldsymbol{v}_k$ in the above equation represent the opening and closing directions of three different groups in a triplet. We sort the scores of all triplets and prioritize them with a score closest to 0 for regrasp planning and uncertainty reduction. The closer the score value is to $0$, the more orthogonal the three grasps from the groups are to each other. By performing successive regrasp using the prioritized triplets, object uncertainties can be effectively minimized. 

It is important to note that strict 90-degree orthogonality is not required to determine the object’s pose. The example shown in Fig. \ref{grasp_planning} represents a special case where the planned grasp poses are right divided into three orthogonal groups with a score of zero. In more general cases, there may be multiple candidate groups and triplet combinations. The triplets may have non-zero scores, However, as long as the grasp direction vectors $\boldsymbol{v}_i$, $\boldsymbol{v}_j$, and $\boldsymbol{v}_k$ form a non-singular matrix when arranged as columns, the pose can be estimated. This point will be clearer in Section V when we introduce the pose estimation algorithm. Here, the reason we sort triplet grasp combinations by their score is not for enforcing strict orthogonality, but rather to account for potential changes of grasp directions caused by subsequent admittance control. Prioritizing grasp triplets with higher orthogonality helps minimize the impact of grasp direction deviations. We will discuss this aspect in more detail in the admittance control section.

\begin{figure}[!htbp]
    \centering
    \includegraphics[width=\linewidth]{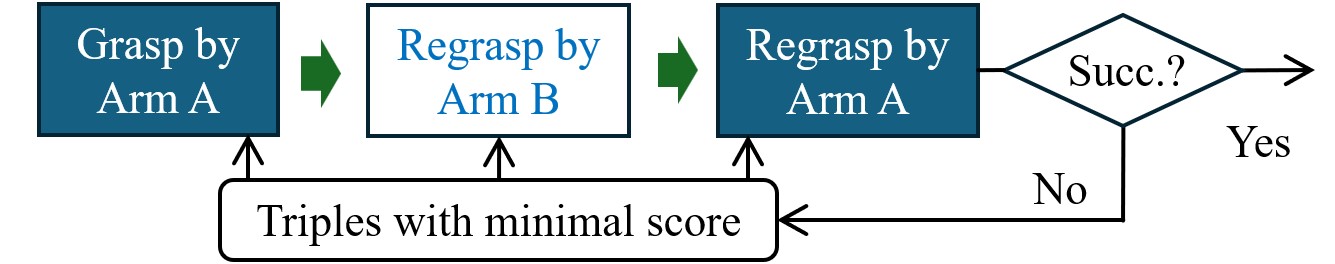}
    \caption{Regrasp planning used in this study is a simple incremental search method, which sequentially traverses grouped grasp triplets until a valid solution is found.}
    \label{planning}
\end{figure}

After identifying the \rev{approximately} orthogonal groups, we leverage the two arms of a bimanual robot to perform regrasp planning interactively for the three grasp groups. We use an incremental method to explore combinations of grasp triplets from the orthogonal groups and output a successful motion once found. Each triplet comprises three groups of grasp poses. During planning, one pose from each group needs to be selected to form a combination of three grasps for computation. In this approach, we do not explicitly extract the three-grasp combinations as a separate step. Instead, we leverage the two arms of a bimanual robot to perform regrasp planning interactively for the three grasp groups. The grasps in each group are sequentially assigned to the two robotic arms for planning. In this way, we incrementally explore the combinations of the grasp triplets and output successful motion once found. The detailed algorithmic flow is shown in Fig. \ref{planning}.

It should be noted that our grasping and planning methods are probabilistic, and their success rates inherently depend on the granularity, number, and diversity of the grasp samples. At the same time, increasing the number of grasps inevitably impacts planning time. Identifying the optimal trade-off between these factors is an important challenge that could be addressed using predictive approaches. If a large amount of data can be accumulated using the proposed method in the future, we are interested in exploring probabilistic modeling methods to learn grasp triplets and thus improve planning efficiency. However, this is beyond the current scope and will not be elaborated.

\section{Conforming to Object Pose Uncertainty}

After planning a feasible sequence of bimanual robot motions, the next step is execution. However, due to uncertainties in the object's pose, a significant discrepancy may exist between the real-world pose and the planning simulation, making pure position control infeasible. This subsection discusses this limitation and highlights the necessity of conforming to object pose uncertainty. It then introduces the admittance control method and explains how it enables the system to adapt to uncertain object poses during handover regrasping.

\begin{figure}[!htbp]
    \centering
    \includegraphics[width=\linewidth]{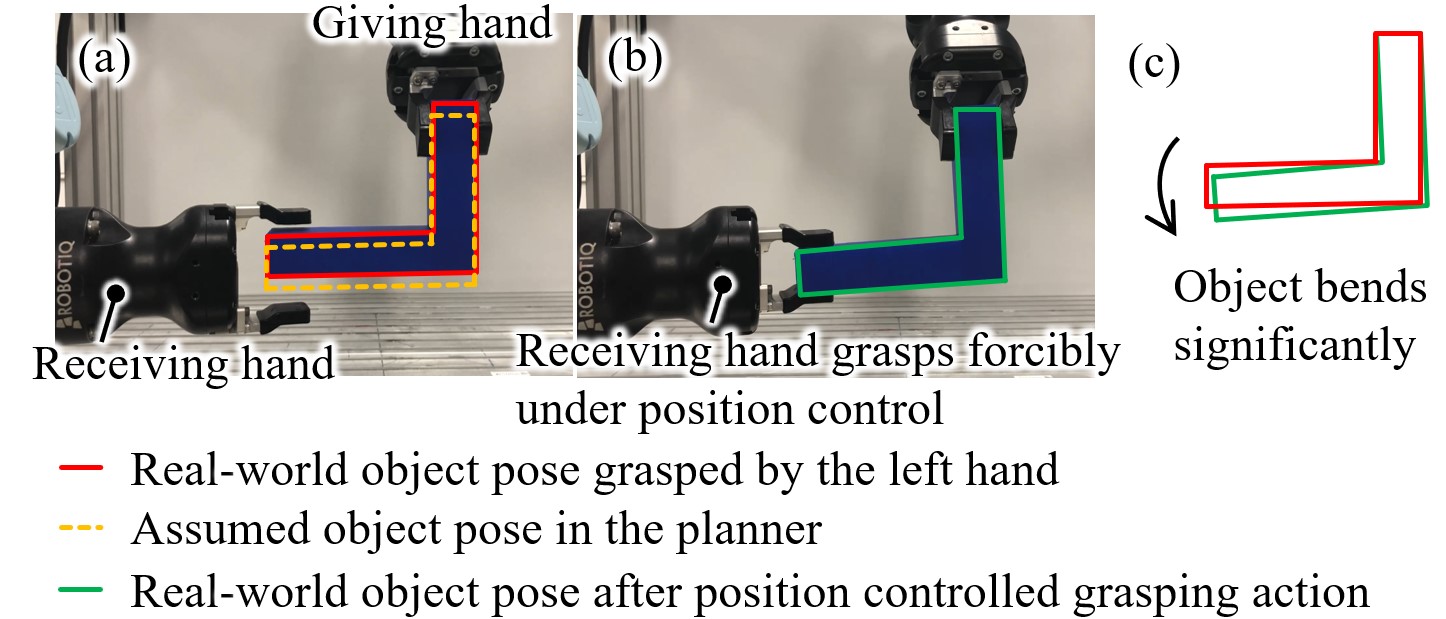}
    \caption{(a) The receiving hand approaches to grasp the object. Due to uncertainties, the object is not positioned exactly at the center of the two fingers, as assumed, but is instead slightly shifted upward. (b) At this point, if position control is used, the upper-side finger will make contact with the object first, exerting excessive force on it and causing deformation. (c) Comparison of object deformation before and after grasping.}
    \label{position_control}
\end{figure}

Consider the scenario shown in Fig. \ref{position_control}(a), where the giving hand has already moved the object from its initial pose to the handover pose, and the receiving hand is approaching to perform the regrasp. The target pose of the giving hand is planned based on an assumed pose in the simulation. However, due to uncertainties in the real world, the object's assumed pose in the simulation deviates from its actual pose, as illustrated by the orange dashed frames and red solid frames in Fig. \ref{position_control}(a). Consequently, the receiving hand's grasping pose becomes misaligned with the actual object pose. Applying position control for grasping in this scenario may lead to undesirable outcomes. In mild cases, as shown in Fig. \ref{position_control}(b) and (c), the object may undergo deformation. In more severe cases, it may sustain damage. Therefore, rigidly executing the planned grasping motion solely with position control is not advisable.

To address this issue, this study employs admittance control during robotic grasping and enables the receiving hand's grasping pose to conform to the uncertain object pose. Admittance control establishes a mass-damper-spring system's motion equation based on the reaction force exerted on the receiving hand and the desired target force \cite{mae2024admittance}. The receiving hand follows the solution of the equation to adjust its motion accordingly, thereby accommodating reaction force. For a desired receiving hand position $\boldsymbol{x}_d$, the admittance control equation is
\begin{equation}
    \boldsymbol{M} (\ddot{\boldsymbol{x}}_p - \ddot{\boldsymbol{x}}_d) + B (\dot{\boldsymbol{x}}_p - \dot{\boldsymbol{x}}_d) + K (\boldsymbol{x}_p - \boldsymbol{x}_d) = \boldsymbol{f} + \boldsymbol{f}_d.
    \label{eqadc}
\end{equation}

Here, $\boldsymbol{x}_p$, $\dot{\boldsymbol{x}_p}$, and $\ddot{\boldsymbol{x}_p}$ are respectively the measured current end-effector position, velocity, and acceleration of the receiving hand. $\boldsymbol{M}$, $\boldsymbol{B}$, and $\boldsymbol{K}$ denote the virtual inertia, damping, and stiffness parameters. $\boldsymbol{f}$ is the reaction force applied to the receiving hand when grasping. It can be measured using the force/torque sensor at the wrist of the receiving manipulator. $\boldsymbol{f}_d$ represents the target exerting force of the receiving hand after applying admittance control. It is essentially zero, as our goal is to ensure that the receiving hand does not exert any external force on the object beyond the necessary grasping force, thereby preventing unintended movement, deformation, or damage. Given the robot's current $\boldsymbol{x}_p$, $\dot{\boldsymbol{x}_p}$, $\ddot{\boldsymbol{x}_p}$, $\boldsymbol{f}$, the desired trajectory $\boldsymbol{x}_d$, $\dot{\boldsymbol{x}}_d$, and $\ddot{\boldsymbol{x}}_d$ can be computed from the admittance control equation. These computed values are then fed to the admittance control law to drive the receiving hand to conform to the object's uncertain pose during grasping.

\begin{figure}[!htbp]
    \centering
    \includegraphics[width=\linewidth]{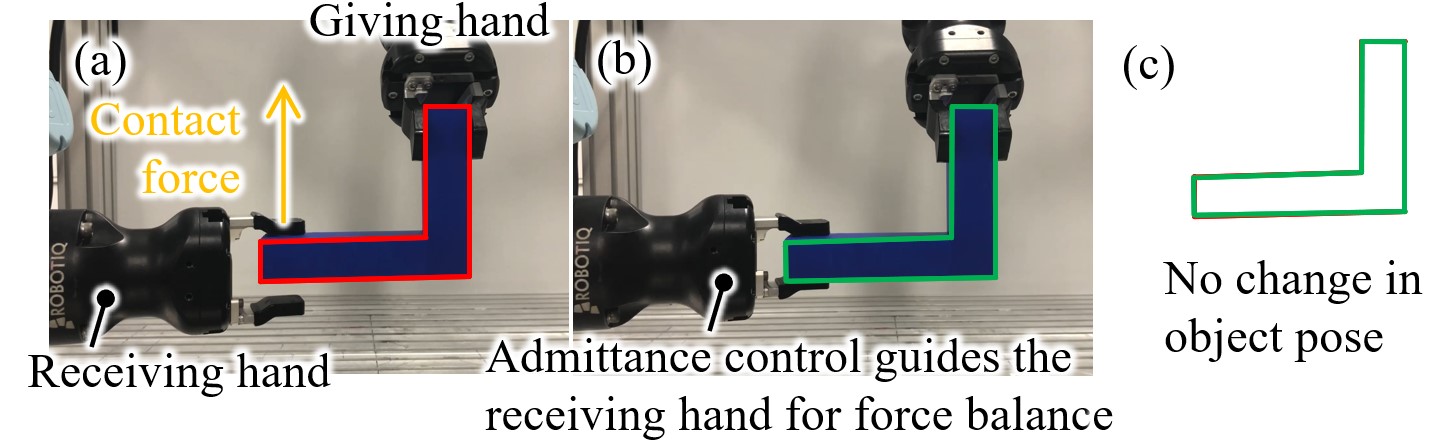}
    \caption{Admittance control helps avoid deformation or damage. (a) A reaction force $\boldsymbol{f}$ is detected when the upper-side finger contacts the object. (b) Admittance control adjusts the receiving hand by moving it to balance $\boldsymbol{f}$. (c) After admittance control, the object is firmly grasped without change.}
    \label{admittance_control}
\end{figure}

Fig. \ref{admittance_control} illustrates the effect of applying admittance control to the grasping process shown in Fig. \ref{position_control}. In Fig. \ref{admittance_control}(a), the upper-side finger first makes contact with the object and experiences a reaction force $\boldsymbol{f}$. The admittance control then adjusts the robotic hand by moving it in a direction that balances this force, as shown in Fig. \ref{admittance_control}(b). The adjustment continues until both fingers are in contact with the object and their grasping forces reach equilibrium. At this point, $\boldsymbol{f}$ disappears, and the grasping forces become internal forces within the whole gripper-object system. The final grasping result is depicted in Fig. \ref{position_control}(c). Unlike position control, admittance control ensures that the object's pose remains unchanged after grasping.

Note that the above equation describes only the position-based admittance control. A similar formulation also applies to rotational admittance control. However, the admittance controller may redirect the grasp toward an adjacent surface due to rotational errors. In extreme cases, this can result in parallel regrasps, making the object's pose indeterminate. As a safeguard against such situations, we suggest prioritizing near-orthogonal grasps following equation \eqref{eq_score}. The design minimizes the likelihood that the three grasps lie on the same plane, even when one of the grasp directions is redirected due to unexpected admittance behavior.

\section{Pose Estimation Using Conformed Grasp Poses}

\subsection{Error Parameterization}

We denote the pose of the object and first grasp in the world frame of the simulation as $({}^w\mathbf{p}_o^\text{sim}, {}^w\mathbf{R}_o^\text{sim})$, and $({}^w\mathbf{p}_{g_1}^\text{sim}, {}^w\mathbf{R}_{g_1}^\text{sim})$, respectively.
The relative pose of the object with respect to the first grasp can be computed as 
\begin{equation}
\left\{
\begin{aligned}
    {}^{g_1}\mathbf{R}_o^\text{sim}& = ({}^w\mathbf{R}_{g_1}^\text{sim})^\top \, {}^w\mathbf{R}_o^\text{sim}\\{}^{g_1}\mathbf{p}_o^\text{sim}& = ({}^w\mathbf{R}_{g_1}^\text{sim})^\top ({}^w\mathbf{p}_o^\text{sim} - {}^w\mathbf{p}_{g_1}^\text{sim})
\end{aligned}\right..
\end{equation}
Meanwhile, we denote the first grasp's actual pose after impedance control as $({}^w\mathbf{p}_{g_1}^{\text{real}}, {}^w\mathbf{R}_{g_1}^{\text{real}})$.
Without considering uncertainty, the pose of the object in the world frame of the real world can be computed as
\begin{equation}
\left\{
\begin{aligned}
{}^w\mathbf{R}_o^\text{ideal}({g_1})& = {}^w\mathbf{R}_{g_1}^{\text{real}} \cdot {}^{g_1}\mathbf{R}_o^\text{sim}\\
{}^w\mathbf{p}_o^\text{ideal}({g_1})& = {}^w\mathbf{p}_{g_1}^{\text{real}} + {}^w\mathbf{R}_{g_1}^{\text{real}} \cdot {}^{g_1}\mathbf{p}_o^\text{sim}
\end{aligned}\right.,
\label{eq_g1est}
\end{equation}
The subscript $(g_1)$ indicates the computations are based on the first grasp. The superscript ``$\text{ideal}$'' indicates the results are an estimation without considering uncertainty.

\begin{figure}[!htbp]
    \centering
    \includegraphics[width=\linewidth]{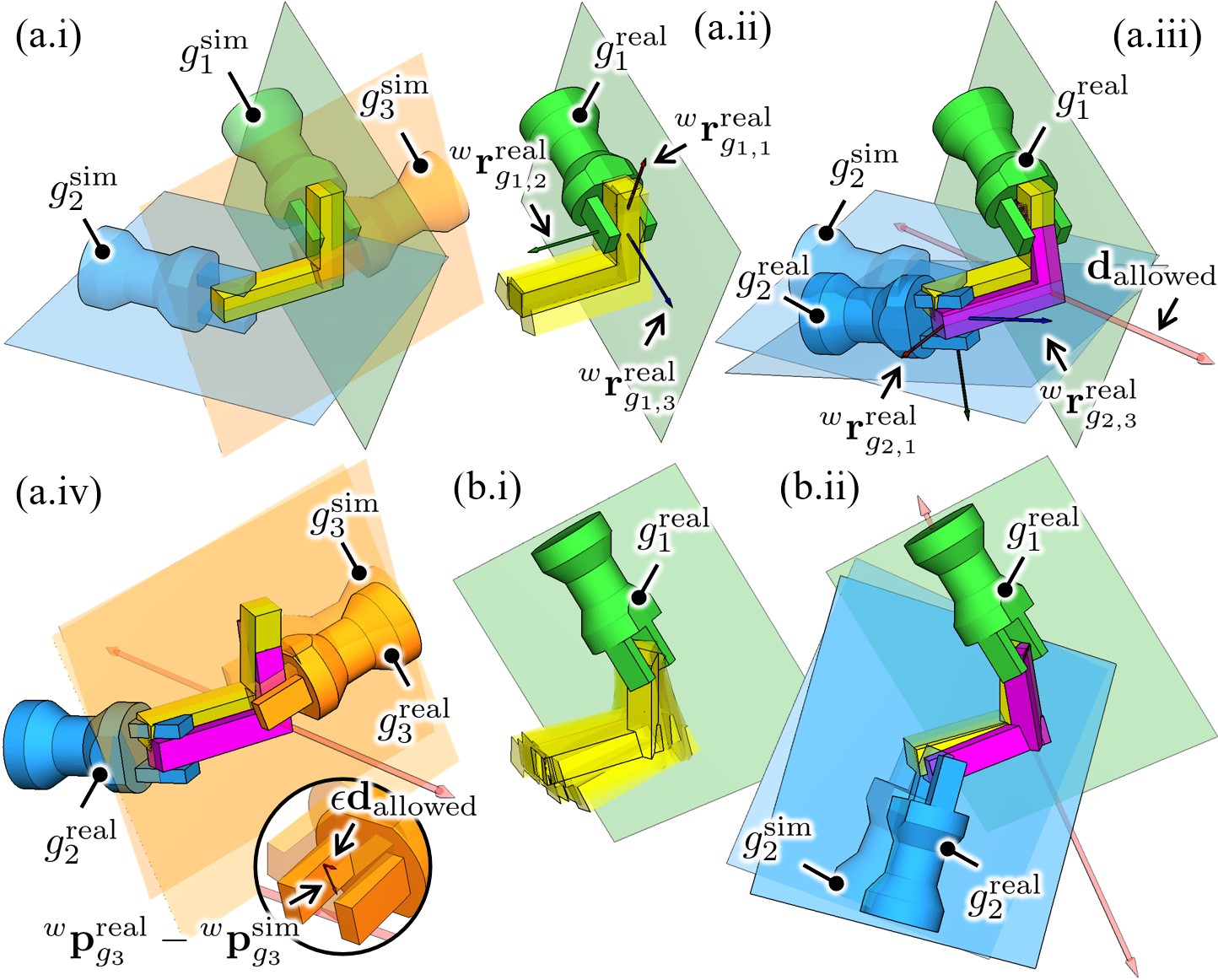}
    \caption{(a) Three grasps of an L-shaped object with a square cross-section (i) and the corresponding constrained uncertainties (ii$\sim$iv). (b) The case of an L-shaped object with a diamond cross-section. The three grasps are not strictly orthogonal. Uncertainty can be estimated as long as they are non-parallel.}
    \label{fig_threeplanes}
\end{figure}

To account for uncertainty introduced during the grasping process, we model the object pose error as comprising a rotational component $\mathbf{R}_o^\text{err}({g_1})$ and a translational component $\mathbf{p}_o^\text{err}({g_1})$. Similarly, $(g_1)$ indicates that the errors are associated with the first grasp. Due to the geometric constraints between the gripper and the object surface, $\mathbf{R}_o^\text{err}({g_1})$ is constrained to a rotation about the second column of the real-world grasp frame, ${}^w\mathbf{r}_{g_{1,2}}^{\text{real}} = {}^w\mathbf{R}_{g_1}^{\text{real}}[:,1]$, and $\mathbf{p}_o^\text{err}({g_1})$ is restricted to the plane spanned by the first and third columns ${}^w\mathbf{r}_{g_{1,1}}^{\text{real}} = {}^w\mathbf{R}_{g_1}^{\text{real}}[:,0]$ and ${}^w\mathbf{r}_{g_{1,3}}^{\text{real}} = {}^w\mathbf{R}_{g_1}^{\text{real}}[:,2]$. The three arrows in Fig. \ref{fig_threeplanes}(a.ii) show these columns. The errors can thus be parameterized as
\begin{equation}
\left\{
\begin{aligned}
\mathbf{R}_o^\text{err}({g_1})& = \exp\left( \theta \, [{}^w\mathbf{r}_{g_{1,2}}^{\text{real}}]_\times \right)\\
\mathbf{p}_o^\text{err}({g_1})& = \delta_{g_{1,1}} \, {}^w\mathbf{r}_{g_{1,1}}^{\text{real}} + \delta_{g_{1,3}} \, {}^w\mathbf{r}_{g_{1,3}}^{\text{real}}
\end{aligned}\right..
\end{equation}
Accordingly, the true pose of the object in the real world is
\begin{equation}
\left\{
\begin{aligned}
{}^w\mathbf{R}_o^{\text{real}}& = {}^w\mathbf{R}_o^{\text{ideal}}(g_1) \cdot \mathbf{R}_o^\text{err}({g_1})\\
{}^w\mathbf{p}_o^{\text{real}}& = {}^w\mathbf{p}_o^{\text{ideal}}(g_1) + \mathbf{p}_o^\text{err}({g_1})
\end{aligned}\right..
\end{equation}

\subsection{Rotation Error Identification}

We can use the second grasp to identify $\mathbf{R}_o^\text{err}({g_1})$. Let $({}^w\mathbf{p}_{g_2}^\text{sim},{}^w\mathbf{R}_{g_2}^\text{sim})$ denote the simulated pose of the second grasp (equally the grasp pose before admittance control), and let $({}^w\mathbf{p}_{g_2}^{\text{real}}, {}^w\mathbf{R}_{g_2}^{\text{real}})$ denote its real-world pose after admittance control. The rotation difference from the simulation to the real execution is $\mathbf{R}_{g_2}^{\text{diff}} = ({}^w\mathbf{R}_{g_2}^\text{sim})^\top \cdot {}^w\mathbf{R}_{g_2}^{\text{real}}$, which captures the rotational deviation introduced by the admittance control. Since $\mathbf{R}_o^\text{err}({g_1})$ is constrained to only occur around $\mathbf{r}_{g_{1,2}}^{\text{real}}$, we extract the component of $\mathbf{R}_{g_2}^{\text{diff}}$ along $\mathbf{r}_{g_{1,2}}^{\text{real}}$ to get $\theta$:
\begin{equation}
\theta = \text{axis\_angle}(\mathbf{R}_{g_2}^\text{err}, {}^w\mathbf{r}_{g_{1,2}}^{\text{real}}),
\end{equation}
and consequently $\mathbf{R}_o^\text{err}({g_1})$. The object's true rotation in the real world can be uniquely identified as
\begin{equation}
{}^w\mathbf{R}_o^{\text{real}} = {}^w\mathbf{R}_o^{\text{ideal}}({g_1}) \cdot \mathbf{R}_o^\text{err}({g_1})={}^w\mathbf{R}_o^{\text{ideal}}({g_1}) \cdot \exp\left( \theta \, [{}^w\mathbf{r}_{g_{1,2}}^{\text{real}}]_\times \right).
\end{equation}
The rotation of the purple object in Fig. \ref{fig_threeplanes}(a.iii) illustrates an identified ${}^w\mathbf{R}_o^{\text{real}}$.

\subsection{Translation Error Identification}

Next, we address $\mathbf{p}_o^\text{err}({g_1})$. In simulation, the pose of the object relative to the second grasp is
\begin{equation}
\left\{
\begin{aligned}
{}^{g_2}\mathbf{R}_o^\text{sim}& = ({}^w\mathbf{R}_{g_2}^\text{sim})^\top \, {}^w\mathbf{R}_o^\text{sim}\\
{}^{g_2}\mathbf{p}_o^\text{sim}& = ({}^w\mathbf{R}_{g_2}^\text{sim})^\top ({}^w\mathbf{p}_o^\text{sim} - {}^w\mathbf{p}_{g_2}^\text{sim})
\end{aligned}\right..
\end{equation}
Let $({}^w\mathbf{p}_{g_2}^{\text{real}}, {}^w\mathbf{R}_{g_2}^{\text{real}})$ be the second grasp's real-world pose. If there is no translational uncertainty, the position of the object in the real world can be computed using the second grasp as
\begin{equation}
    {}^w\mathbf{p}_o^{\text{ideal}}({g_2}) = {}^w\mathbf{p}_{g_2}^{\text{real}} + {}^w\mathbf{R}_{g_2}^{\text{real}} \, {}^{g_2}\mathbf{p}_o^\prime.
    \label{eq_g2est}
\end{equation}
Since ${}^w\mathbf{R}_o^{\text{real}}$ has already been identified, the corrected relative position ${}^{g_2}\mathbf{p}_o^\prime$ instead of ${}^{g_2}\mathbf{p}_o^\text{sim}$ is used in the equation. ${}^{g_2}\mathbf{p}_o^\prime$ is computed as
\begin{equation}
    {}^{g_2}\mathbf{p}_o' = {}^{g_2}\mathbf{R}_o^{\text{real}} \left( {}^{g_2}\mathbf{R}_o^\text{sim} \right)^\top {}^{g_2}\mathbf{p}_o^\text{sim}, 
\end{equation}
where
\begin{equation}
    {}^{g_2}\mathbf{R}_o^{\text{real}} = ({}^w\mathbf{R}_{g_2}^{\text{real}})^\top \, {}^w\mathbf{R}_o^{\text{real}}.
\end{equation}

The translational uncertainty caused by the second grasp lies within the plane spanned by the first and third columns of ${}^w\mathbf{R}_{g_2}^{\text{real}}$ and can be parameterized as 
\begin{equation}
\mathbf{p}^\text{err}({g_2}) = \delta_{2,1} \, {}^w\mathbf{r}_{g_{2,1}}^{\text{real}} + \delta_{2,3} \, {}^w\mathbf{r}_{g_{2,3}}^{\text{real}},
\end{equation}
where ${}^w\mathbf{r}_{g_{2,1}}^{\text{real}} = {}^w\mathbf{R}_{g_2}^{\text{real}}[:,0]$ and ${}^w\mathbf{r}_{g_{2,3}}^{\text{real}} = {}^w\mathbf{R}_{g_2}^{\text{real}}[:,2]$. Fig. \ref{fig_threeplanes}(a.iii) illustrates them. The true position of the object can be expressed using $\mathbf{p}^\text{err}({g_2})$ as
\begin{equation}
{}^w\mathbf{p}_o^{\text{real}} = {}^w\mathbf{p}_o^{\text{ideal}}(g_2) + \mathbf{p}_o^\text{err}({g_2}).
\end{equation}

It is important to note that the translational errors associated with the first and second grasps are respectively constrained to planes determined by $({}^w\mathbf{r}_{g_{1,1}}^{\text{real}},{}^w\mathbf{r}_{g_{1,3}}^{\text{real}})$ and $({}^w\mathbf{r}_{g_{2,1}}^{\text{real}},{}^w\mathbf{r}_{g_{2,3}}^{\text{real}})$. Their normal vectors are $\mathbf{n}_{g_1} = {}^w\mathbf{r}_{g_{1,1}}^{\text{real}} \times {}^w\mathbf{r}_{g_{1,3}}^{\text{real}}$ and $\mathbf{n}_{g_2} = {}^w\mathbf{r}_{g_{2,1}}^{\text{real}} \times {}^w\mathbf{r}_{g_{2,3}}^{\text{real}}$. Since the translational error must simultaneously satisfy the constraints imposed by both planes, it is restricted to lie along their intersection line. The direction of this line can be computed using
\begin{equation}
\mathbf{d}_\text{allowed} = \frac{\mathbf{n}_{g_1} \times \mathbf{n}_{g_2}}{\|\mathbf{n}_{g_1} \times \mathbf{n}_{g_2}\|}.
\end{equation}
Thus, the translational error can essentially be parameterized using a single scalar value as $\epsilon\,\mathbf{d}_\text{allowed}$, and the true real-world object position simplifies to
\begin{equation}
{}^w\mathbf{p}_o^{\text{real}} = {}^w\mathbf{p}_o^{\text{ideal}}({g_2}) + \epsilon \, \mathbf{d}_\text{allowed}.
\label{eq_poserr}
\end{equation}

The scalar value $\epsilon$ can be easily determined from the displacement of the third grasp after admittance control. Let ${}^w\mathbf{p}_{g_3}^\text{sim}$ and ${}^w\mathbf{p}_{g_3}^\text{real}$ denote the simulated position of the third grasp (position before admittance control) and the real-world position after admittance control, respectively. The displacement is computed as ${}^w\mathbf{p}_{g_3}^\text{real} - {}^w\mathbf{p}_{g_3}^\text{sim}$. The scalar $\epsilon$ can be obtained using the following projection
\begin{equation}
\epsilon = \mathbf{d}_\text{allowed}^\top ({}^w\mathbf{p}_{g_3}^\text{real} - {}^w\mathbf{p}_{g_3}^\text{sim})),
\end{equation}
as indicated within the circle in Fig. \ref{fig_threeplanes}(a.iv). Substituting the obtained $\epsilon$ into equation \eqref{eq_poserr} yields the object's true position.

In addition to the L-shaped object with a square cross-section discussed thus far, Fig. \ref{fig_threeplanes}(b) presents an L-shaped object with a diamond cross-section. Although this object does not contain any pair of strictly orthogonal grasping directions, it is still possible to select three non-parallel grasps and estimate the final object pose using the method described in this section. Fig. \ref{fig_threeplanes}(b.i) shows the uncertainty after the first grasp, while Fig. \ref{fig_threeplanes}(b.ii) illustrates the orientation determined after the second grasp, along with the direction of the remaining translational error. A detailed 3D visualization of both the L-shaped objects can be found in the supplementary video.

\section{Experiments and Analysis}

\begin{figure}[!tbp]
    \centering
    \includegraphics[width=\linewidth]{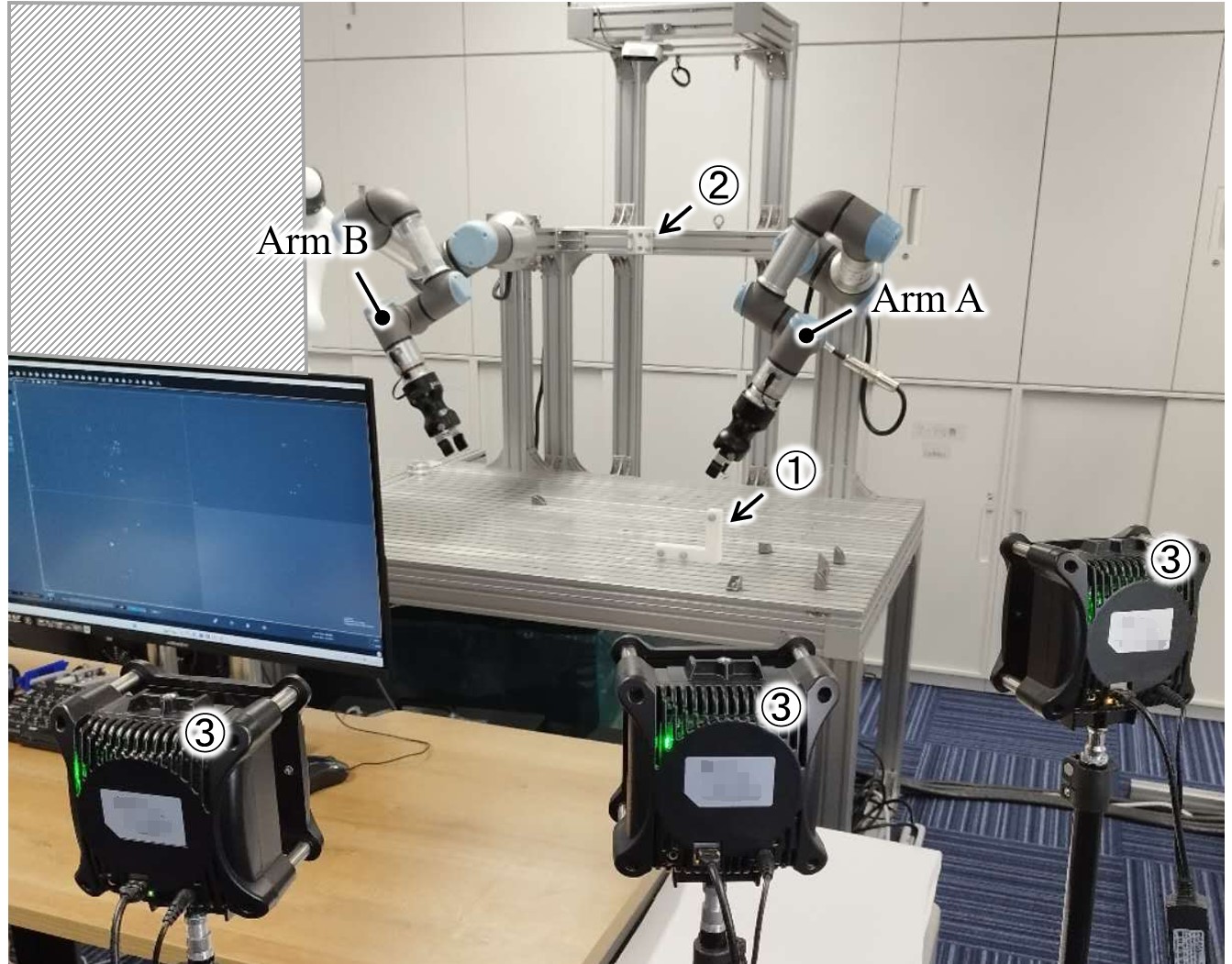}
    \caption{Bimanual platform used in the experiments. \circnum{1} Target object with optical markers. \circnum{2} Markers for global reference. \circnum{3} Cameras used for detecting optical markers.}
    \label{exp_env}
\end{figure}

Our experimental platform consists of two UR3e robots. Their setup is shown in Fig. \ref{exp_env}. Each robot is installed with a Robotiq Hand-E gripper. The opening range of the gripper is 0$\sim$50 mm. The regrasp planning is performed using the WRS simulation environment. The coordinate system of the platform is defined at the back of the robot, with its origin at the intersection of the central plane between the two robot arms and the ground. The $\boldsymbol{x}$-axis points toward the monitor in the figure. The $\boldsymbol{y}$-axis points towards the first arm. The $\boldsymbol{z}$-axis points upward. For comparison, we attached optical markers to both the objects and the robots, and tracked the poses of the resulting rigid bodies using three tracking cameras. The poses estimated by the proposed method were compared with those captured by the optical tracking system to evaluate the method’s performance. The tracking system provides a detection accuracy of $\pm$0.5 mm.

\begin{figure}[!tbp]
    \centering
    \includegraphics[width=\linewidth]{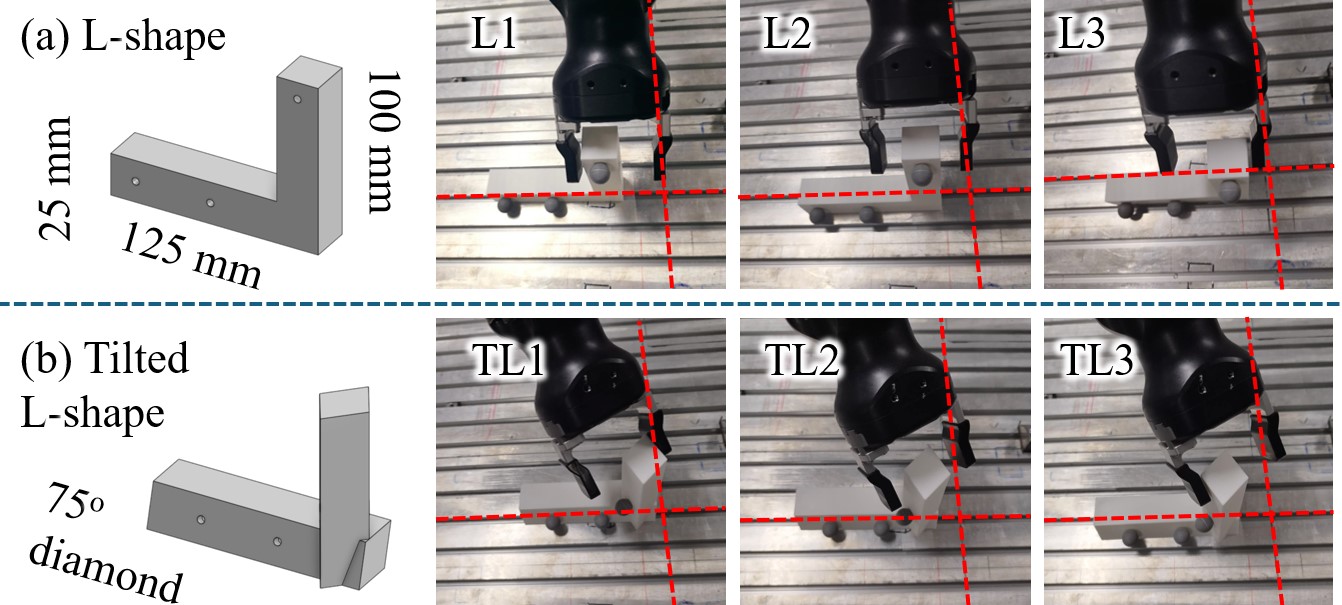}
    \caption{(a) L-shape object and its random initial placements. (b) TL-shape object and its random initial placements.}
    \label{exp_trials}
\end{figure}

Fig. \ref{exp_trials} shows the objects used in the experiments, their dimensions, and the randomly assigned placements in experimental trials. The objects are L-shaped items with square and diamond cross-sections, as described earlier. Their long arm measures 125 mm, the short arm 100 mm, and the distance between the parallel edges of the cross-section is 25 mm for both. The diamond-shaped cross-section has a 75${}^\circ$ acute interior angle. Optical tracking markers were attached at pre-drilled holes on the objects. During grasp planning, we can automatically generate grasps and motions that avoid collisions with the markers. For each object, we randomly assigned three different initial placements, labeled L1$\sim$3 and TL1$\sim$3 in Fig. \ref{exp_trials}, and tasked the robots with moving the objects to predefined target positions. The supplementary video accompanying this paper contains details of the grasp selection, motion planning, and robot actions. We recommend readers to refer to this video for an intuitive understanding of the experimental procedures.

Table \ref{tab_diff} shows the results. The data section has two main column groups reporting the positional differences and rotational differences. Within each group, there are three subcolumns. The \rev{``second grasp''} columns show the positional and rotational differences of the $g_2$ grasp before and after admittance control. The \rev{``third grasp''} columns present the corresponding differences for the $g_3$ grasp. The ``object'' columns report the differences between the object pose estimated by the proposed method and the pose measured by the optical tracking system. \rev{The ``first grasp'' uses position control and is thus excluded.}

\begin{table*}[ht]
\setlength{\tabcolsep}{5pt}
\centering
\caption{Experimental Results with Statistics}
\label{tab_diff}
\begin{threeparttable}
\begin{tabular}{c|
ccc|ccc|>{\columncolor{orange!20}}c>{\columncolor{orange!20}}c>{\columncolor{orange!20}}c|
ccc|ccc|>{\columncolor{orange!20}}c>{\columncolor{orange!20}}c>{\columncolor{orange!20}}c}
\toprule
\multirow{2.5}{*}{} & 
\multicolumn{9}{c|}{{Position Differences} ($\Delta \mathbf{p}$, mm)} &
\multicolumn{9}{c}{{Rotation Vector Differences} ($\Delta \mathbf{w}$, degrees)} \\
\cmidrule(lr){2-10} \cmidrule(lr){11-19}
&
\multicolumn{3}{c|}{\rev{second grasp}} &
\multicolumn{3}{c|}{\rev{third grasp}} &
\multicolumn{3}{>{\columncolor{orange!20}}c|}{object} &
\multicolumn{3}{c|}{\rev{second grasp}} &
\multicolumn{3}{c|}{\rev{third grasp}} &
\multicolumn{3}{>{\columncolor{orange!20}}c}{object} \\
\midrule
L1 &
\phantom{-}0.40 & -1.60 & 5.20 &
-8.90 & \phantom{-}2.00 & -0.80 &
-33.70 & -61.80 & 32.20 &
\phantom{-}0.55 & \phantom{-}0.91 & \phantom{-}0.02 &
\phantom{-}0.34 & \phantom{-}0.75 & -0.91 &
\phantom{-}0.01 & 2.82 & -1.39 \\
L2 &
\phantom{-}0.30 & -1.70 & 5.20 &
\phantom{-}0.20 & \phantom{-}2.00 & -0.80 &
-33.60 & -61.70 & 32.30 &
\phantom{-}0.51 & \phantom{-}0.14 & \phantom{-}0.03 &
\phantom{-}0.33 & -0.25 & \phantom{-}0.06 &
\phantom{-}0.12 & 1.83 & -1.95 \\
L3 &
\phantom{-}0.40 & -1.60 & 5.20 &
\phantom{-}9.00 & \phantom{-}1.90 & -0.40 &
-33.20 & -61.50 & 31.70 &
\phantom{-}0.30 & \phantom{-}0.13 & \phantom{-}0.00 &
\phantom{-}0.20 & -0.51 & \phantom{-}1.11 &
-0.00 & 0.98 & -1.07 \\
\midrule
Mean &
-- & -- & -- & -- & -- & -- &
-33.50 & -61.67 & 32.07 &
-- & -- & -- & -- & -- & -- &
\phantom{-}0.04 & 1.87 & -1.47 \\
Std  &
-- & -- & -- & -- & -- & -- &
\cellcolor{lime!70}\phantom{-0}0.25 & \cellcolor{lime!70}\phantom{-0}0.15 & \cellcolor{lime!70}\phantom{0}0.25 &
-- & -- & -- & -- & -- & -- &
\cellcolor{lime!70}\phantom{-}0.06 & \cellcolor{lime!70}0.95 & \cellcolor{lime!70}\phantom{-}0.44 \\
\midrule
TL1 &
-1.81 & \phantom{-}0.99 & 0.38 &
\phantom{-}8.02 & \phantom{-}8.05 & -7.20 &
-59.35 & \phantom{-}32.75 & 52.41 &
\phantom{-}0.64 & \phantom{-}1.07 & -1.00 &
\phantom{-}0.45 & \phantom{-}0.36 & \phantom{-}0.78 &
-0.01 & 2.05 & \phantom{-}0.42 \\
TL2 &
-0.08 & -0.18 & 1.31 &
\phantom{-}0.40 & -1.60 & 5.20 &
-59.66 & \phantom{-}34.42 & 51.76 &
-0.00 & -0.00 & \phantom{-}0.00 &
\phantom{-}0.30 & \phantom{-}0.13 & \phantom{-}0.00 &
\phantom{-}0.13 & 0.73 & \phantom{-}1.09 \\
TL3 &
\phantom{-}0.07 & -0.21 & 1.17 &
-1.71 & -0.36 & -1.47 &
-56.37 & \phantom{-}36.34 & 52.26 &
-0.04 & -0.01 & -0.00 &
\phantom{-}1.25 & -0.29 & \phantom{-}0.56 &
\phantom{-}0.86 & 0.27 & \phantom{-}2.11 \\
\midrule
Mean &
-- & -- & -- & -- & -- & -- &
-58.46 & \phantom{-}34.50 & 52.14 &
-- & -- & -- & -- & -- & -- &
\phantom{-}0.33 & 1.02 & \phantom{-}1.21 \\
Std  &
-- & -- & -- & -- & -- & -- &
\cellcolor{lime!70}\phantom{-0}1.86 & \cellcolor{lime!70}\phantom{-0}1.50 & \cellcolor{lime!70}\phantom{0}0.36 &
-- & -- & -- & -- & -- & -- &
\cellcolor{lime!70}\phantom{-}0.47 & \cellcolor{lime!70}0.38 & \cellcolor{lime!70}\phantom{-}0.85 \\
\bottomrule
\end{tabular}
\end{threeparttable}
\end{table*}

From the table, we observe that both the $g_2$ and $g_3$ grasps exhibit substantial changes before and after admittance control, indicating that the robot actively adjusts the regrasp pose to estimate uncertainties for each task. However, the final estimation results are not satisfactory, showing positional differences of up to 61.67 mm and rotational differences of up to 1.86 degrees compared to the optical tracking system.

Upon closer examination, we found that these errors are primarily systematic. For the L-shaped object, the standard deviations in $\Delta \mathbf{p}$ across tasks were [0.25, 0.15, 0.25] mm, all below 0.3 mm, and the deviations in $\Delta \mathbf{w}$ were [0.06, 0.95, 0.44] degrees, all below 1.0 degrees. Similarly, for the TL-shaped object, the deviations in $\Delta \mathbf{p}$ and $\Delta \mathbf{w}$ were [1.86, 1.50, 0.36] mm and [0.47, 0.38, 0.85] degrees, respectively, remaining below 2 mm and 0.9 degrees. These results suggest that while the overall estimation errors appear large, their internal consistency is high, indicating that the system’s inaccuracies stem largely from systematic biases. \rev{Likely sources include calibration errors in the optical tracking system, installation errors in the two robot arms and the mounting aluminum frame, and absolute control errors of the robots.} In addition, when comparing the deviations between the L-shaped and TL-shaped objects, we found that the inference results for the L-shaped object were better than those for the TL-shaped object. This outcome differs from our theoretical expectations. \rev{We can conclude from the equations in Section V that even if the three grasps are not perfectly orthogonal, the final inference accuracy should not degrade, and the L- and TL-shaped objects should exhibit similar deviations.} We believe that the observed difference arises because the TL-shaped object involves rotations around axes unaligned with body frames. The large difference indicates that the robot setup has greater rotational misalignment. \rev{This raises the question of compensating for system bias to improve absolute accuracy. While theoretically feasible, such correction would require complex estimation of both positional and rotational errors, which we leave for future work.}

In conclusion, although the results exhibit large absolute errors, the deviation levels are on the same order of magnitude as the optical tracking system. The results indicate that the proposed method has strong relative pose estimation performance.

\section{Conclusions and Future Work} 
This paper proposes a method for reducing object uncertainties by leveraging the grasping constraints of parallel grippers with flat finger pads and a bimanual regrasp planning method. We applied this method to estimate grasping errors for a roughly placed L-shaped object. The results demonstrate that the proposed approach can effectively reduce uncertainties and achieves high repeatability accuracy.

\rev{In our experiments, we used a dual-arm setup built from two single-arm robots mounted on an aluminum frame, which inevitably introduced systematic errors and led to relatively large absolute errors. As a next step, we plan to validate the method in a dynamic simulation environment, where no absolute error is present, and use a dual-arm robot with commercially-calibrated body integration for practical deployment.}

\normalem
\bibliographystyle{IEEEtranN}
\bibliography{citations.bib}

\end{document}